\documentclass[conference]{IEEEtran}
\IEEEoverridecommandlockouts
\usepackage{cite}
\usepackage{csquotes}
\usepackage{amsmath,amssymb,amsfonts}
\usepackage{algorithmic}
\usepackage{longtable, adjustbox}
\usepackage{graphicx}
\usepackage{textcomp}
\usepackage{xcolor}
\usepackage{color, colortbl}
\usepackage{hyperref}
\definecolor{Gray}{gray}{0.9}

\usepackage{float}
\usepackage{array}
\newcolumntype{L}{>{\centering\arraybackslash}m{3cm}}
\def\BibTeX{{\rm B\kern-.05em{\sc i\kern-.025em b}\kern-.08em
    T\kern-.1667em\lower.7ex\hbox{E}\kern-.125emX}}

\makeatletter
\newcommand{\linebreakand}{%
  \end{@IEEEauthorhalign}
  \hfill\mbox{}\par
  \mbox{}\hfill\begin{@IEEEauthorhalign}
}
\makeatother

\author{
  \IEEEauthorblockN{Suriyadeepan Ramamoorthy}
\IEEEauthorblockA{\textit{School of Electronic Engineering} \\
\textit{Dublin City University}\\
Dublin, Ireland \\
suriyadeepan.ramamoorthy2@mail.dcu.ie}
  \and
  \IEEEauthorblockN{Joyce Mahon}
\IEEEauthorblockA{\textit{School of Computer Science} \\
\textit{University College Dublin}\\
Dublin, Ireland \\
joyce.mahon1@ucdconnect.ie}
  \and
  \IEEEauthorblockN{Michael O’Mahony}
\IEEEauthorblockA{\textit{School of Computer Science} \\
\textit{Technological University Dublin}\\
Dublin, Ireland \\
michael.t.omahony@mytudublin.ie}
  \linebreakand 
  \IEEEauthorblockN{Jean Francois Itangayenda}
\IEEEauthorblockA{\textit{School of Info. \& Commun. Studies} \\
\textit{University College Dublin}\\
Dublin, Ireland \\
jean.itangayenda@ucdconnect.ie }
  \and
  \IEEEauthorblockN{Tendai Mukande}
\IEEEauthorblockA{\textit{School of Electronic Engineering} \\
\textit{Dublin City University}\\
Dublin, Ireland \\
tendai.mukande2@mail.dcu.ie}
  \and
  \IEEEauthorblockN{Tlamelo Makati}
\IEEEauthorblockA{\textit{School of Computer Science} \\
\textit{Technological University Dublin}\\
Dublin, Ireland \\
D21124389@mytudublin.ie }
}

\title{Automatic Contact Tracing using Bluetooth Low Energy Signals and IMU Sensor Readings}
\begin{document}
\maketitle
\begin{abstract}

In this report, we present our solution to the challenge provided by the SFI Centre for Machine Learning (ML-Labs) in which the distance between two phones needs to be estimated. It is a modified version of the NIST Too Close For Too Long (TC4TL) Challenge, as the time aspect is excluded. 
We propose a feature-based approach based on Bluetooth RSSI and IMU sensory data, that outperforms the previous state of the art by a significant margin, reducing the error down to $0.071$. 
We perform an ablation study of our model that reveals interesting insights about the relationship between the distance and the Bluetooth RSSI readings..

\end{abstract}

\begin{IEEEkeywords}
Contact Tracing, Bluetooth, IMU, RSSI, TC4TL
\end{IEEEkeywords}
\section{Introduction}
\label{intro}
The Coronavirus (COVID-19) pandemic has  profoundly impacted the world, with effects ranging from food insecurity and extreme poverty to widespread fatalities. Digital Contact Tracing has emerged as a cost-effective means of harnessing  technologies such as GPS, QR codes, Wi-Fi and Bluetooth to trace and notify users of their interactions with potentially infected persons; and automated tracing apps deployed on smartphones may be used to identify close contacts. Bluetooth is the most widely used digital contact tracing technology due to its low energy consumption and cost. \cite{gasser} \cite{shahroz} \\
\indent In 2020, the Too Close for Too Long (TC4TL) Challenge  \cite{evaluation_plan} was organized by the National Institute of Standards and Technology (NIST) in partnership with the MIT Private Automated Contact Tracing (PACT) research group. The TC4TL Challenge is concerned with proximity sensing - the aim is to predict whether two individuals have been in \enquote{close contact} for \enquote{too long}. This challenge involves estimating the distance and time between two phones given a series of received signal strength indicator (RSSI) values along with  inertial measurement unit (IMU) sensory data.   The initial task was later modified so as to only focus on the distance (too close) element \cite{evaluation_plan}.
A number of teams \cite{LCD} \cite{contact_tracing_group} \cite{contact_tracing_group_ppt} participated in the challenge, and their machine learning (ML) model performance scores are recorded on an evaluation leader board on the NIST TC4TL Challenge website. \cite{tc4tl} Later in 2020, groups from the SFI  Centre  for  Machine Learning (ML-LABS) participated in a challenge that also focused on the distance (too close)  element only (using the same TC4TL Challenge criteria and data). Gómez et al \cite{gomez} developed two ML models -  Gradient Boosting Machine (GBM) and Multi-layer Perceptron (MLP) trained on simple features generated from Bluetooth Low Energy (BLE) RSSI readings and experiment conditions, that surpassed the rest of the models in the NIST leader board.
These results are discussed later in this report, and also refered to in Table \ref{table:results}.
\indent{}Estimating distance between devices using BLE RSSI readings is a non-trivial task.
Signal strengths are easily influenced by a number of factors other than the distance between devices.
These factors include shadowing effect due to obstruction between the transmitter and the receiver that cause scattering and reflection during transmission, angle between the devices, the properties of radios used in devices, etc.
To overcome these challenges, an effective model should make use of the IMU sensor readings along with the information provided about the experiment conditions. 
The next challenge lies in how to make use of the IMU sensor readings.
IMU sensors such as accelerometer and gyroscope are highly sensitive to noise and not a reliable estimator of distance on their own.
Moreover they typically vary across x, y, z dimensions and are observed over a period of time.
This raises an interesting question. Should we manually craft features to reduce these high-dimensional observations to meaningful predictors of distance or should we rely on Deep Neural Networks to automatically learn smooth complex functions that map these observations to distance classes? This report describes how we have taken on this modified challenge, and built a series of models using the given data, with the aim of improving the accuracy of estimating the distance between two phones. A key question being answered in this report is how to make effective use of IMU sensor readings to compensate for the environmental factors that Bluetooth signals are sensitive to.
    
    
    
    


\section{Challenge Description}
The data used in this challenge was provided through the NIST portal \cite{tc4tl} and it contained 15,552 training files, 8,423 test files, and 936 development files. Each contact event file represented the data collected between a transmitter and a receiver device over a  period of time. The development and test sets were extracted from the MIT Matrix Data data set \cite{matrix_github}, while the training set was extracted from the MITRE Range-Angle Structured data set\cite{mitre_github} \cite{sajadi}. Both data sets were collected using the Structured Contact Tracing Protocol \cite{protocol}.

Each event file consists of a header containing information about the experiment conditions, such as the position of the device, types of devices used, etc,. This is followed by a series of sensor readings and BLE signal strength readings with timestamps. The frequency of the recordings varies from event to event. During the data collection process, the readings were measured in windows of 4 seconds, called chirps or looks. Each event has a step size variable (10s, 20s, 30s, ..., 150s) associated with it, that indicate how frequently these looks were performed. Data collection is designed in such a way that the distance is fixed during each look, and monotonically increasing on every step through the event \cite{sajadi} \cite{protocol}. In addition to the event files, key files are provided separately for training, development and test sets, which contain the step size and distance $d$ (between devices) information. There are two subsets in the data sets - fine-grain and coarse-grain distances; and $d$ can vary within a single contact event file by up to 0.9 m and 2.1 m, respectively. Each event file consists of multiple 4 second looks worth of BLE RSSI readings and IMU sensor readings, temporally separated based on the event-specific step size. The objective of the challenge is to map each event to a distance class $\in \{1.2, 1.8, 3., 4.5\}$. 

Outputs of the models were originally submitted through the NIST scoring software, and the performance of the model were determined by evaluating the probability of miss and the probability of false alarm. The two probabilities were combined through the use of a normalised decision cost function (nDCF). The challenge weights have been set by NIST \cite{tc4tl}. The test set and its key file, along with a submission scoring software are provided to the participants for evaluating their models locally. The NIST scoring portal has been closed during the time of this work.

\section{Related Work}

Distance estimation and indoor localization using received signal strength are well explored problems in Wireless networks \cite{wn1} \cite{wn2}.
Contact Tracing using distance automatically estimated using BLE signal strengths is a relative new problem that gained traction due to the advent of the covid-19 pandemic. As mentioned in Section \ref{intro}, RSSI is a noisy indicator of distance as it is dependent upon many other factors. Xenakis et al \cite{xenakis} identifies hardware (radio) and firmware specifications among major factors that influence RSSI. BLE accommodates 40 channels within the 2402 MHz - 2480 MHz band. Each channel exhibits different properties resulting in difference in behaviour of transmitted signal. They experimented with different orientation combinations between two devices and noted unpredictable variations in signals strengths.

Researchers have tried different approaches to compensate for these factors. He et al \cite{contact_tracing_group} use BLE RSSI histogram as a feature to reduce signal fluctuation due to multi-path effect. They also use IMU sensor readings to compensate for signal blocking from phone carriage states. Shankar et al \cite{shankar} have identified that using only the bluetooth RSSI data for modeling results in overfitting that can be alleviated by introducing IMU sensor readings producing a regularization effect.

The challenge of high dimensionality of IMU sensor data had been addressed by Shankar et al \cite{shankar} by modeling the problem as a time-series classification task. They constructed a feature vector at every time step containing all the sensor readings along with BLE RSSI values at that instant. This series of feature vectors representing an event, is mapped to a distance class using a recurrent neural network. They have also experimented with several fixed-length models, both neural network-based and tree-based. They have reported an nDCF score of $0.16$ achieved by a temporal one-dimensional convolutional neural network using GRU gating mechanism (ConvGRU). Others have bypassed the curse of dimensionality by reducing the high dimensional sensory data into a few useful scalar features. He et al \cite{contact_tracing_group} have engineered features such as BLE histogram for noise-invariance and carriage state extracted from linear acceleration which in turn was estimated using IMU sensory data. They build a vanilla neural network to learn a mapping between these features and distance classes resulting in an average nDCF score of $0.58$ \cite{contact_tracing_group_ppt}. Similarly, Gomez et al \cite{gomez} rely on radio propagation-based features including a linear approximation model of distance and RSSI and path loss attenuation. They also use normalized Bluetooth RSSI averaged over an event as a feature, along with experiment conditions such as device type, pose, etc,. A GBM trained on these features achieves an average nDCF score of $0.5175$.

\section{Methodology}

\begin{figure}[h]
    \centering
    \includegraphics[width=0.45\textwidth]{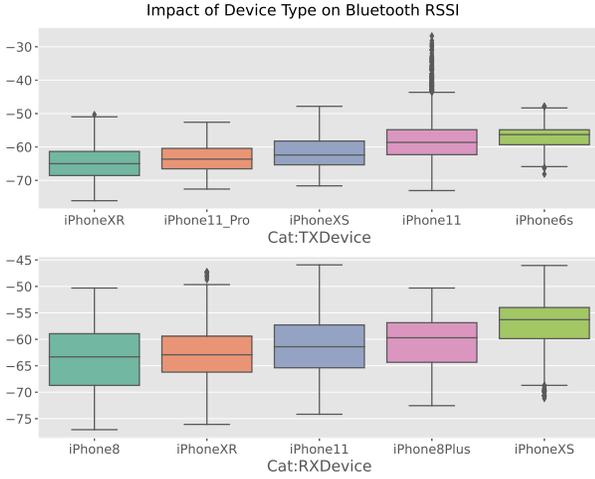}
    \caption{Bivariate Device on BLE}
    \label{fig:bivariate_device}
\end{figure}

\begin{figure}[h]
    \centering
    \includegraphics[width=0.45\textwidth]{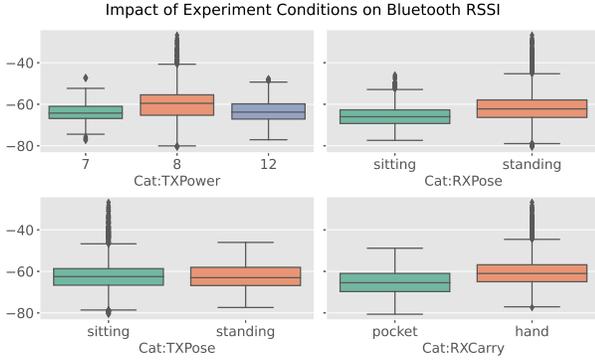}
    \caption{Bivariate Power Pose Carry}
    \label{fig:bivariate_power}
\end{figure}

We have conducted extensive Exploratory Data Analysis in search for variables that are correlated with the target variable, the distance.
Correlation study showed that while the bluetooth RSSI values are strongly correlated with distance, there is no significant correlation between distance and the experiment conditions, like device type, pose, etc.
We investigated the influence of these categorical variables on the magnitude of bluetooth RSSI.
The results presented in Figure \ref{fig:bivariate_device} and Figure \ref{fig:bivariate_power} indicate a strong correlation between the experiment conditions and the bluetooth signal strength.
So we have included these variables as features in all our models in order to compensate for their influence over the signal strength.

In all our experiments, we have learned hyperparameters associated with the model and the training process, using sklearn's grid search method \cite{sklearn}. We train our models on the training set, learn hyperparameters and conduct model selection by evaluating on the development set. We conduct a dual evaluation by splitting up the datasets into coarse-grained and fine-grained subsets, and use two instances of classifiers for training and evaluation. To calculate the nDCF score, we concatenate the predictions from the coarse-grained and fine-grained models and input them to the submission scoring software. The final nDCF scores are calculated by training the selected model on the training and development set concatenated together and evaluated on the test set.

\subsection{Simplistic Baseline}
Unlike other related works, Gomez et al \cite{gomez} proved to be an attractive candidate for the baseline due to its simplicity. 
The features include a linear approximation of the relationship between distance and RSSI given by equation \ref{eq:lam}.
The series of Bluetooth RSSI values spread across time steps was summarized into a scalar by taking average over the whole event.
Path Loss Attenuation was calculated using equation \ref{eq:pla} and normalized. 
Apart from these continuous features, experimental conditions including device carry position, device pose, type of device were encoded as categorical features using ordinal encoding.
We were able to reproduce the results of Gomez et al, by training a GBM on these features.
Experiments were conducted with different combinations of these features and results are presented overleaf in Figure \ref{fig:gomez_ablation}.


\begin{equation} \label{eq:lam}
\textbf{d'} =10^{\frac{TX - RSSI}{10*N}}
\end{equation}

\begin{equation} \label{eq:pla}
P_{L} = P_{t} - 41 - RSSI
\end{equation}



\subsection{Radio Propogation Modeling}
Similar to Gomez et al \cite{gomez}, we used radio propagation models to estimate distance between devices as a function of Bluetooth RSSI. The parameters of the model are learned by running a Bayesian search over the parameter space that tries to minimize the error $(d - d')$. There are a number of radio propagation models available in Wireless networks literature that account for different factors affecting the signal strength. We experimented with two simple models - linear approximation model (equation \ref{eq:lam}) and Friis Free Space Propagation model \cite{rpmodels}. The Friis model is used to model propagation path loss incurred in a free-space environment, devoid of objects that cause absorption, diffraction or reflections. The Friis equation for received signal strength, in log scale is given by:
\begin{equation}
    P_r = P_t + G_t + G_r + 20 log_{10}(\lambda) - 20 log(4 \pi d) - 10 log (L)
\end{equation}

where $P_t$ is the power of the transmitted signal (dBm), $G_t$ \& $G_r$ are transmitter and receiver antenna gains (dBi) correspondingly, $\lambda$ is the wavelength of the carrier signal and $L$ represents other losses not associated with the propagation loss. All the parameters are learned similarly through a Bayesian search over the training set. We have chosen hyperopt\cite{hyperopt}, a simple, flexible hyperparameter search framework for this purpose.



\subsection{Bluetooth and Beyond}
We improved over our baseline, by increasing the number of features used to represent the Bluetooth RSSI readings for an event.
Gomez et al \cite{gomez} drastically reduced the series of RSSI observations into one scalar by taking an average over them.
We instead summarized the bluetooth readings using multiple summary statistics including the minimum, mean and maximum. 
This resulted in a slight reduction in the nDCF error measure.
Following this trend, we added several percentile values of RSSI observations to enhance the summary, leading to a richer representation.
Feature selection was conducted to learn a set of percentile features that resulted in the lowest nDCF error.

The experiment conditions provided a rich context to each event.
As suggested in He et al \cite{contact_tracing_group} and Gomez et al \cite{gomez}, we used this information by encoding them as categorical variables.
This resulted in a drastic reduction in nDCF error. 
To further improve the performance of our model, we added simple summaries (min, mean, max) of IMU sensor readings, resulting in a further improvement over the previous score, with average nDCF being reduced to $0.071$, - see Table \ref{table:results}


\begin{table} [!h]
\centering
\resizebox{0.5\textwidth}{!}{
    \begin{tabular}{|p{2.5cm} | p{0.7cm} p{0.7cm}p{0.7cm}p{0.7cm}||p{0.7cm}|}
    \hline
    \textbf{} & Fine Grain 1.2m & Fine Grain 1.8m & Fine Grain 3.0m & Coarse Grain 1.8m & Average nDCF \\ 
    
    \hline
    Extra Trees Classifier & \text{0.078} & \text{0.061} & \text{0.070} & \text{0.076} & \textbf{0.071} \\

     \hline
    Radio Prop. Model & \text{0.524} & \text{0.504} & \text{0.574} & \text{0.358} & \textbf{0.489} \\
   
   \hline
    Shankar et al \cite{shankar} & \text{-} & \text{-} & \text{-} & \text{-} & \text{0.18} \\
    
    \hline
    Gomez et al \cite{gomez}  & \text{0.6} & \text{0.52} & \text{0.58} & \text{0.37} & \text{0.5175} \\

    \hline
    He et al \cite{contact_tracing_group}  & \text{0.68} & \text{0.54} & \text{0.59} & \text{0.41} & \text{0.555} \\
    
    \hline
    Team: LCD \cite{LCD}
  & \text{0.6} & \text{0.58} & \text{0.63} & \text{0.55} & \text{0.59} \\
    \hline
    \end{tabular}} \vspace{0.25cm}
    \caption{Our results compared with related works}
\label{table:results}
\end{table}

\section{Experimental Results \& Discussion}
We used sklearn's GBM ensemble to model the mapping between the features proposed in Gomez et al \cite{gomez} and the distance classes. The hyperparameters of GBM were learned by running a grid search on the development set. Their final results of Gomez et al were successfully reproduced by training the GBM on the training and the development sets concatenated together and evaluated on the test set. We achieved an nDCF score of $0.52$ which is close enough to the original results ($0.5175$). An ablation study was conducted to understand the significance of individual features.

\begin{figure}[h]
    \centering
    \includegraphics[width=0.5\textwidth]{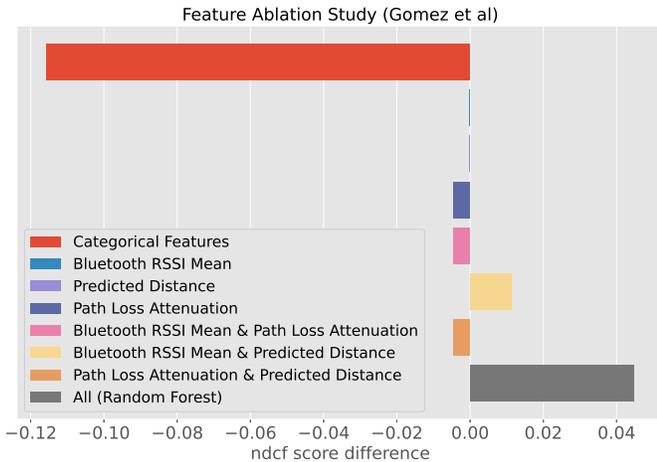}
    \caption{Ablation study on Gómez et al \cite{gomez}}
    \label{fig:gomez_ablation}
\end{figure}

Features were removed from the baseline methodically and evaluation was performed on models trained on the rest of the features.
The results of this study are presented overleaf in Figure \ref{fig:gomez_ablation}. Each bar in Figure \ref{fig:gomez_ablation} represents the change in performance of the baseline when a particular feature indicated by the label associated with the bar, is removed from the model. A decrease along the x-axis indicates the negative effect of the removal of the feature, while an increase indicates a positive effect of removal of the feature.
It is obvious from the results that the categorical features representing the experiment conditions are crucial for distance estimation.
It is also evident that the three numerical features - Bluetooth RSSI mean, Path loss attenuation and Predicted Distance are interchangeable. 
In fact, Bluetooth RSSI mean and Predicted Distance from the list of features, improves the model performance, reducing the nDCF score by 0.01. 
We have evaluated several classifier ensembles against the test set and report that the Random Forest ensemble outperforms the GBM.

\begin{figure}[h]
    \centering
    \includegraphics[width=0.40\textwidth]{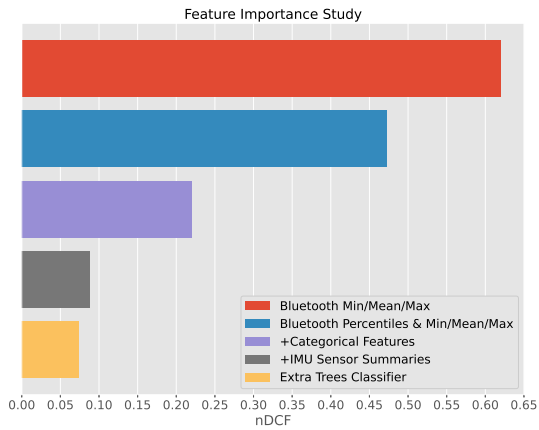}
    \caption{Feature Importance Study on our Proposed Model}
    \label{fig:ram_importance}
\end{figure}

We have evaluated Linear Approximation model and Friis Free Space model of radio propagation along with Bluetooth summaries.
The results of the experiments presented in Table \ref{table:radio}, indicate that neither of the models are useful.
Neither of them capture any information more than that is already present in the bluetooth summaries.
We hypothesize that without placing a strong prior on the parameters of the radio propagation models, nothing useful can be learned from the data.
Such strong priors on parameters such as antenna gain, should come from domain expertise. Using insights gathered from the ablation study of baseline features and the performance evaluation of radio propagation models, we learned that the contribution of  Bluetooth RSSI
readings in its various forms to the model performance is limited.
\begin{table} [h!]
\begin{center}
\begin{tabular}{|p{6.9cm} || p{0.7cm}|}
    \hline
    &\\
\textbf{Features} & \textbf{nDCF}\\
\hline
Bluetooth Min/Mean/Max & 0.619\\
\hline 
Bluetooth Percentiles & 0.473\\
\hline 
Linear Approximation Model with Bluetooth Min/Mean/Max & 0.615\\ 
\hline 
Linear Approximation Model with Bluetooth Percentiles & 0.489\\
\hline 
Fris Free Space Model with Bluetooth Min/Mean/Max & 0.615\\ 
\hline 
Fris Free Space Model with Bluetooth Percentiles & 0.499\\
\hline 
\end{tabular}
\vspace{0.25cm}
\caption{Performance of Radio Propagation Models}
\end{center}
\label{table:radio}
\end{table}
In order to improve the results, we needed additional features that are unrelated to Bluetooth sensor readings. Bluetooth statistical summaries including min/mean/max and percentiles, lead to a nDCF score of $0.47$. Adding categorical features improved the performance by a significant margin, resulting in $nDCF = 0.21$. We ran a search over all the categorical encoders available in the \emph{category\_encoders} \cite{ce} python package, resulting in Polynomial Coding being the best performing encoding mechanism.
By applying the same summarization technique to IMU sensor readings, we engineered features of type \emph{Sensor:Min}, \emph{Sensor:Mean}, and \emph{Sensor:Max}. Including these features produced the best performing model by far reducing the average nDCF score down to a shocking $0.090$, beating the previous state of the art score of $0.18$ \cite{shankar}. To improve the performance further, we evaluated multiple classifier ensembles and ended up with Extra Trees Classifier outperforming the rest, reducing the score even further down to $0.071$. These results are visually presented in Figure \ref{fig:ram_importance}.

\begin{figure}[h]
    \centering
    \includegraphics[width=0.5\textwidth]{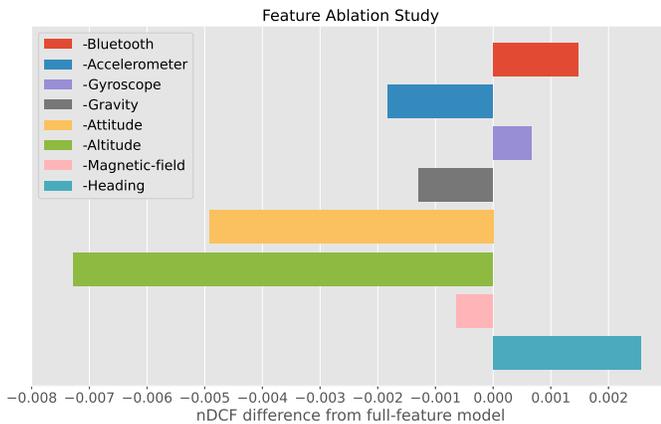}
    \caption{Ablation Study on our Proposed Model}
    \label{fig:ram_ablation}
\end{figure}

An Ablation study was conducted on the final model by removing features methodically and evaluating the model with the remaining features.
The change in performance of the model is represented by the bar chart in Figure \ref{fig:ram_ablation}.
The first bar denotes an increase in performance when the Bluetooth sensory features were removed.
Even if the increase of $0.0015$ is considered insignificant, we still end up with a highly accurate model which doesn't rely on Bluetooth sensory data.
This observation disproved our assumption that the Bluetooth RSSI readings were the central feature for distance estimation. 
Although \emph{Altitude} might seem like feature critical to the model performance, it should be noted that a reduction of $0.007$ is practically insignificant. We conclude that the sensory zones of the IMU sensor units overlap with each other considerably, resulting in interchangeable features.
We also hypothesize that the distance between devices is encoded in some form, within the IMU sensor readings. We leave it to the researchers reading our work to investigate this phenomenon and draw their conclusions.

\section{Future Work}
Even though our model performs exceedingly well, we were unable to exactly model the effect of individual variables on distance or their effect on each other. We invite researchers to investigate the effect of experiment conditions on the BLE signal strength and consider building a translation function that maps BLE signal strength profile of device A to device B. The impact of orientation on signal strength needs to be studied using observations from gyroscope and accelerometer. Mathematically modeling the relationship between distance and RSSI, improving over existing radio propagation models like Friis Free Space model or Log-normal shadowing model \cite{rpmodels}, by adding one or more compensation terms derived from IMU sensor readings is another challenge we leave to the future.

\section{Ethical Considerations}
Any consideration of the implementation of a contact tracing app arising in the context of the COVID-19 pandemic must also bear in mind the wider significance of such a decision. 
\subsection{The Irish and European Context}

Ní Fhaoláin et al \cite{nifhaolain} assessed the appetite for trustworthiness and the regulation of artificial intelligence (AI) in Europe, by examining the National Strategies of the 29 signatories of the 2018 Declaration of Cooperation on AI \cite{declaration}. This study found that in general, these signatories highlighted AI ethics and regulation as an area that requires attention; and "there is an expectation that Signatories themselves will attempt to regulate AI or that they will join other countries to regulate". The European Commission’s Coordinated Plan on AI: 2021 Review. \cite{review} identified a set of joint actions for the European Commission and Member States: 1) set enabling conditions for AI development and uptake in the EU; 2) make the EU the place where excellence thrives from the lab to the market; 3) ensure that AI works for people and is a force for good in society; and 4) build strategic leadership in high-impact sectors. These actions are reflected in the 2021, National AI Strategy for Ireland.  \cite{gov}

\subsection{Potential Issues in the Implementation of AI}

The Trinity College Dublin funded “Testing Apps for COVID-19 Tracing” (TACT) project, highlights the issue of false positive proximity warnings in contact tracing apps, so as to disconcert people or to discredit the system. They believe that: 1) preventing a bad actor attack could add significant complexity to the overall system and might not be feasible; 2) that the impact of the attack increases as more people run the tracing app; and 3) that the attack can be targeted against key staff in some scenarios so that targeting even with a small amplification factor may cause noticeable damage \cite{farrell}. Leith and Farrell identified that  many health authorities depend upon the non open source Google/Apple Exposure Notification (GAEN) API for their operation. This has limited public documentation, and uses a filtered BLE signal strength measurement that can be potentially misleading with regard to the proximity between two handsets \cite{leith}.
 
\subsection{Additional Societal Considerations}

These include: 1) an ethical assessment should be analysed against the scale of the deaths and suffering; 2) the justification of blanket lockdowns would be weaker were it possible to manage physical distancing in a more evidence-based, risk-adjusted way; 3) there are privacy concerns relating to data collection scope and duration; 4) an app has the potential to be autonomy enhancing; 5) the use of incentives could adversely impact those who do not have access to suitable smart-phones; 6) an app could provide a way for professionals and institutions to meet their obligations;  7) the public availability of source code, supporting material, etc; 8) the reproducibility and consistency of results \cite{gasser} \cite{vinuesa} \cite{stephens} .
\section*{Conclusion}
In this paper, the approach for detecting proximity between phones for automatic contact tracing is presented.
We propose a feature engineering approach that relies on the statistical summaries of Bluetooth RSSI readings, IMU sensory data and experiment conditions encoded as categorical variables.
Our model, an Extra Trees Classifier ensemble trained on these features, outperforms the state of art by a significant margin, reducing the average nDCF score to $0.071$.
Ablation study conducted on our model reveals the interchangeable nature of both IMU sensors and Bluetooth RSSI observations for estimating distance, subverting the popular opinion that Bluetooth RSSI readings are the best predictor of distance.
We theorize that the distance between devices is encoded in some form in the IMU sensor observations.

\section*{Acknowledgment}
This publication has emanated from research conducted with the financial
support of Science Foundation Ireland under Grant number 18/CRT/6183. For the purpose
of Open Access, the author has applied a CC BY public copyright licence to any
Author Accepted Manuscript version arising from this submission. We would also like to acknowledge Mayug Maniparambil for mentoring us; and Omid Sadjadi, for providing us with information about the data.


\end{document}